\documentclass[10pt, a4paper]{article}

\usepackage{lrec-coling2024} % this is the new style
\usepackage{times}
\usepackage{latexsym}
\usepackage{amsmath}
\usepackage{amsfonts}
\usepackage{comment}
\usepackage[T1]{fontenc}
\usepackage[utf8]{inputenc}

\usepackage{microtype}
\usepackage[ruled,vlined]{algorithm2e}
\usepackage{multirow}
\usepackage{amsmath, xparse}
\usepackage{makecell}
\usepackage{ragged2e}
\usepackage{xcolor}
\usepackage{graphicx}
\usepackage{subfigure}
\usepackage{booktabs,caption}
\usepackage[flushleft]{threeparttable}

\usepackage{inconsolata}

\title{\normalfont{  \textbf{Few-Shot Relation Extraction with Hybrid Visual Evidence}}}

\name{Jiaying Gong, Hoda Eldardiry} 

\address{Virginia Tech\\
         Blacksburg, VA, USA\\
         \{gjiaying, hdardiry\}@vt.edu\\}

\abstract{
The goal of few-shot relation extraction is to predict relations between name entities in a sentence when only a few labeled instances are available for training. 
Existing few-shot relation extraction methods focus on uni-modal information such as text only. This reduces performance when there is no clear contexts between the name entities described in text.
We propose a multi-modal few-shot relation extraction model (MFS-HVE) that leverages both textual and visual semantic information to learn a multi-modal representation jointly. The MFS-HVE includes semantic feature extractors and multi-modal fusion components. The MFS-HVE semantic feature extractors are developed to extract both textual and visual features. The visual features include global image features and local object features within the image.
The MFS-HVE multi-modal fusion unit integrates information from various modalities using image-guided attention, object-guided attention, and hybrid feature attention to fully capture the semantic interaction between visual regions of images and relevant texts.       
Extensive experiments conducted on two public datasets demonstrate that semantic visual information significantly improves performance of few-shot relation prediction. 
 \\ \newline  }

\begin{document}

\maketitleabstract
\section{Introduction}

Relation extraction aims to predict the relation between two name entities in a sentence.
To alleviate the reliance on high-quality annotated data, few-shot learning has drawn more attention, requiring only a few labeled instances for training to adapt to new tasks.
Existing few-shot relation extraction methods can be roughly divided into two categories.
One category involves methods only using plain text data, without any auxiliary information.
For example, meta-learning models prototypical networks~\cite{Gao_Han_Liu_Sun_2019}, siamese neural networks~\cite{8258168} are trained with only a few examples for each class to extract relations.
The other category introduces external data sources such as relation information~\cite{liu-etal-2022-learn, liu-etal-2022-simple}, concepts of entities~\cite{yang-etal-2021-entity}, side information~\cite{10.1145/3459637.3482403}, external datasets~\cite{10.1145/3340531.3411858}, and graphs~\cite{qu2020few}, to compensate the limited information in the above methods, to enhance the performance in few-shot relation extraction.

\begin{figure}[htp] 
 \center{\includegraphics[height=3cm,width=7.5cm]{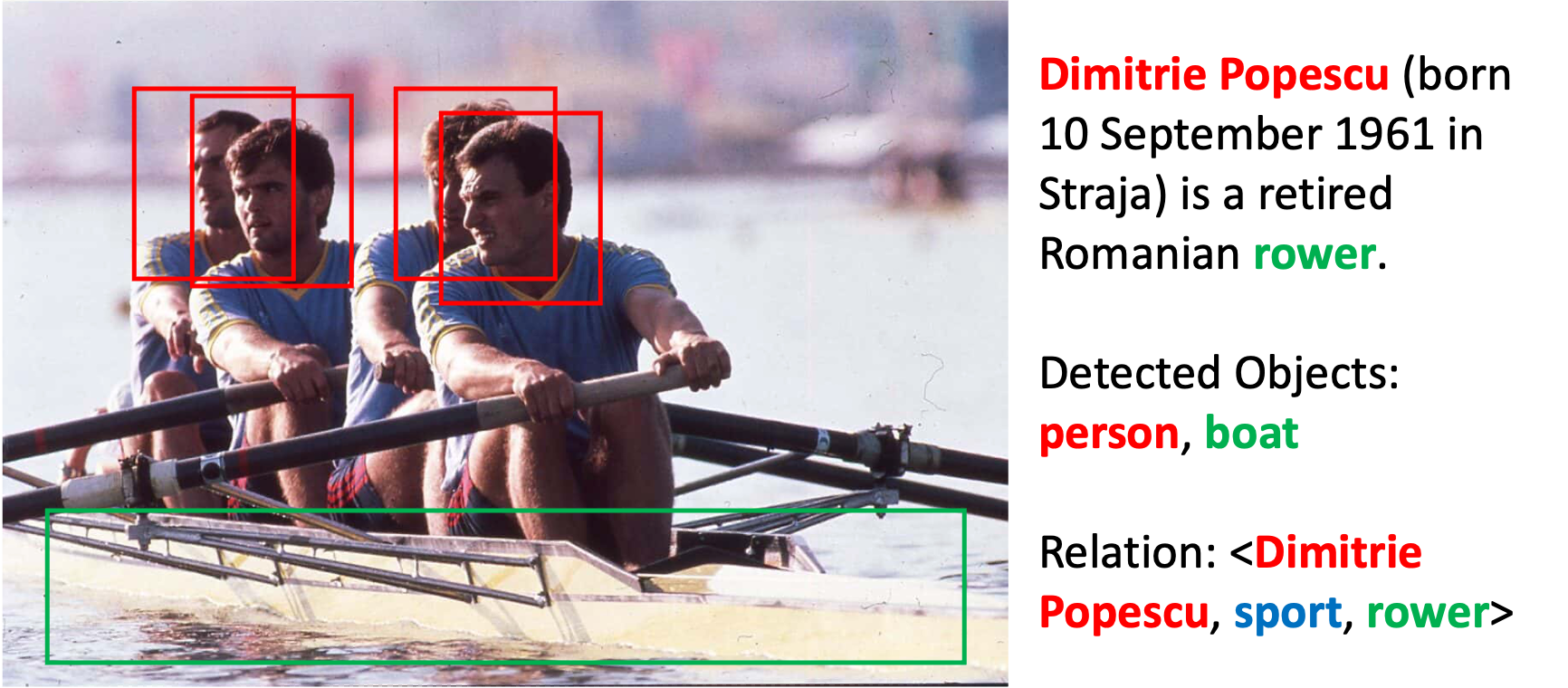}}
 \caption{\label{fig:example} An example of multi-modal relation extraction based on visual information.}
 \end{figure} 

However, these methods mainly explore single-modality text-based data and may suffer a significant performance decline when texts lack contexts. 
For example, in Figure~\ref{fig:example}, given two name entities `Dimitrie Popescu' and `rower', it is difficult for text-based models to detect the relation `sport' without other supplementary information because the word `sport' or other similar words does not appear in the text.
As a result, uni-modal models will incorrectly extract the relation `winner' or `candidate' of the two name entities according to the short given textual sentence. 
Even models using external information such as knowledge graphs or related words with similar meanings still can not correctly extract the relation due to the limited information in short given textual sentences.

Therefore, we question that \textit{Can visual information be a good external source to supplement the missing contexts in textual sentences for few-shot relation extraction?}
In the above case, we can easily classify the relation into `winner' from the guidance of an image showing that a person is holding a trophy. 
Utilizing visual information to support contextual information for texts involves multi-modal learning.
However, fusing information from different modalities is also a challenging task.
First, simply concatenating textual and visual features without considering semantic information may even have a negative impact on the performance as shown in Sec.~\ref{sec:experiment}.
For example, in Figure~\ref{fig:example}, the multiple people's faces in the background are noise for the image with the relation `sport'.
Second, existing multi-modal models (Sec.~\ref{sec:related}) mainly focus on fusing global visual features with text without considering the semantic information of visual objects in images.
In Figure~\ref{fig:example}, visual objects such as `person' and `boat' contain essential information to the relation `sport'.

To address these challenges, we propose a \textbf{M}ultimodal \textbf{F}ew-\textbf{S}hot model based on \textbf{H}ybrid \textbf{V}isual \textbf{E}vidence (MFS-HVE) for relation extraction. 
We first generate the representations through the textual feature extractor in Sec.~\ref{sec:textExtractor} and the visual feature extractor in Sec.~\ref{sec:viExtractor}. 
We consider the visual representations from both the local perspective in low resolution (Sec.~\ref{sec:obAtt}) and the global perspective in high resolution (Sec.~\ref{sec:piAtt}).
To be more specific, a local feature vector is the embedding of the objects detected from the image, and a global feature vector is the embedding of the whole image.
Because local features only focus on objects, global features can overcome the problem of sparsity with more information; however, they may probably contain noise (irrelevant information).
We integrate both local features and global features to solve the problem of sparsity and noise.

Secondly, inspired by the cross-modal attention mechanism~\cite{10.1145/3404835.3462924}, we propose a multi-modal fusion unit including image-guided attention, object-guided attention, and hybrid feature attention to integrating semantic information from different modalities at both global and local levels.
%We adopt cross-modal guided attention to integrate semantics with different modalities to fully connect visual and textual information at both global and local levels.
From the global perspective, image-guided attention based on the scaled dot-product attention~\cite{NIPS2017_3f5ee243} combines global feature vectors from the image with texts to capture the semantic interaction between visual regions of images and texts.
From the local perspective, object-guided attention fuses objects detected from the image with relevant name entities from the textual sentences.
Then the hybrid feature attention fuses all textual and visual information, including global image features and local object features.
The hybrid feature attention generates a weight vector, multiplied by the multi-modal representations.

Finally, we concatenate text features, image-guided features, and object-guided features through a cross-modality encoder to generate the final multi-modal representations.
% based on the prototypical networks~\cite{NIPS2017_cb8da676}.
Each relation representation is calculated based on the prototypical networks~\cite{NIPS2017_cb8da676}.
Next, based on the prototypical networks~\cite{NIPS2017_cb8da676}, we compute the mean value of all multi-modal support vectors as the prototype to represent each relation.
%Next, we represent the multi-modal relation based on the prototypical networks~\cite{NIPS2017_cb8da676}.
Because of the hierarchical structure of the detected objects and name entities discussed in Sec.~\ref{sec:feaAtt}, hyperbolic distance is calculated between multi-modal query representations and prototypes to predict the relation.
We conduct extensive experiments on two public datasets MNRE~\cite{10.1145/3474085.3476968} and FewRel~\cite{han-etal-2018-fewrel} to evaluate whether semantic visual information can supplement the missing contexts in textual sentences for few-shot relation extraction.
FewRel is a uni-modal dataset containing only text, we crawl the image automatically by icrawler~\footnote{https://icrawler.readthedocs.io/en/latest/} for each instance to provide visual information, which can facilitate future research on multi-modal few-shot relation extraction.
Details are introduced in Sec.~\ref{sec:dataset}.% and Appendix~\ref{sec:A1}.
By comparing MFS-HVE with some state-of-the-art uni-modal few-shot relation extraction models and some multi-modal fusion methods with the same feature extractors, we show that, in general, models with multi-modal information perform better than the text-only models.
%However, our experimental results show that simply fusing all information directly without considering semantic contexts may have a negative impact.
%Our proposed model MFS-HVE, which integrates multi-modal semantic information at both global and local levels with three different attentions, outperforms other SOTA multi-modal fusion techniques introduced in Sec.~\ref{sec:baselines}.
We also conduct ablation studies and parameter sensitivity studies to learn the impact of each attention and function.
The contributions of this paper can be summarized as:
\begin{itemize}
    \item We propose the first approach (MFS-HVE) for multi-modal few-shot relation extraction. Existing models for few-shot relation extraction only focus on a single data modality.
    \item MFS-HVE combines information from different modalities through image-guided attention, object-guided attention, and hybrid feature attention to integrating semantic visual information and textual information. %at both global and local levels.
    \item We conduct extensive experiments on two public datasets. The experimental results show that introducing visual information can supplement the missing contexts in textual sentences for the few-shot relation extraction task. 
\end{itemize}

\begin{figure*}[htp] 
 \center{\includegraphics[height=5cm,width=\textwidth]{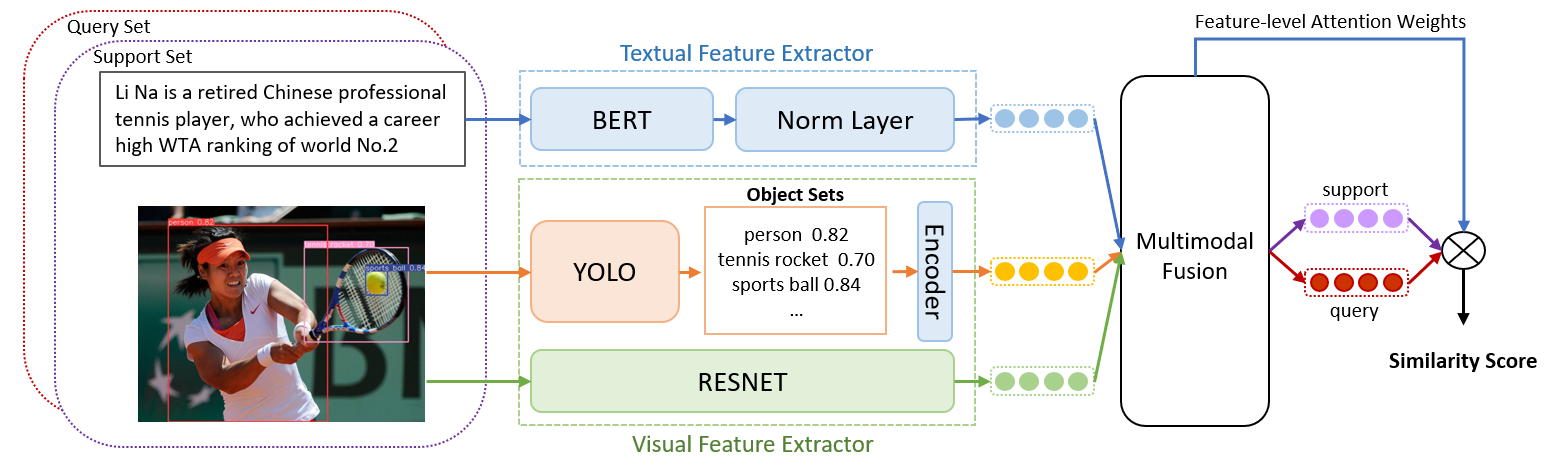}}
 \caption{\label{fig:overview} The overview of MFS-HVE. Details of multi-modal fusion is introduced in Sec.~\ref{sec:fusion} and Figure~\ref{fig:fusion}}
 \end{figure*} 
 
\section{Related Work}%Done
\subsection{Few-shot Relation Extraction}
%Relation extraction predicts the relation of two name entities expressed in a sentence.
Recent studies of few-shot relation extraction focused on metric-based representative methods.
For example, the prototypical network learns a prototype for each relation via instance embeddings~\cite{Gao_Han_Liu_Sun_2019, ye-ling-2019-multi, baldini-soares-etal-2019-matching, few-shot2020}.
Siamese neural network learns the metric of relational similarities between pairs of instances~\cite{8258168, Gao2020NeuralSF}.
Additional data sources are also used to help improve the performance in few-shot learning.
Meta information such as relation information~\cite{dong-etal-2020-meta, liu-etal-2022-learn, liu-etal-2022-simple, zhenzhen-etal-2022-improving, ijcai2022p407, li-qian-2022-graph, zhang-lu-2022-better}, concepts of entities~\cite{DBLP:conf/coling/WangBLWHZZ20, yang-etal-2021-entity}, additional auxiliary information~\cite{10.1145/3459637.3482403}, knowledge from cross domains~\cite{10.1145/3340531.3411858}, data augmentation~\cite{qin-joty-2022-continual, gong2023promptbased}, and global graphs of all relations~\cite{qu2020few} are considered as prior information to establish connections between instance-based information and conceptual semantic-based information.
However, the above studies only explore uni-modal text data. 
Different from these studies, we propose utilizing different data modalities, including both textual information and visual information, to supplement the missing semantics in texts.

\subsection{Few-Shot Multi-Modal Fusion}~\label{sec:related}
Few-shot multi-modal fusion extracts relevant information from different modalities and integrates information collaboratively.
%Multi-modal learning has been well studied in the areas of visual question answering~\cite{singh-etal-2021-mimoqa, Chappuis_2022_CVPR}, name entity recognition~\cite{10.1145/3394171.3413650, 10.1145/3488560.3498475}, and most related relation extraction tasks in supervised learning~\cite{9428274, 10.1145/3474085.3476968, chen-etal-2022-good}.
%These methods have demonstrated that the performance of these tasks can be improved by fusing information from different modalities.
%Inspired by these works, we consider fusing visual information for few-shot relation extraction to provide the missing context in texts.
%Recent works~\cite{Wan_Zhang_Du_Huang_Yang_Pan_2021}
%They miss the semantic information of visual objects in images.
%Multi-modal fusion include concatenation of features, tensor transformations~\cite{ijcai2018-143}, attention modules~\cite{Osman2019DRAUDR, Yu_2019_CVPR} and cross-modal graph alignment~\cite{10.1145/3474085.3476968, wang-etal-2022-ita} to integrate features from different modalities.
MNRE is the first dataset developed for multimodal relation extraction~\cite{zheng2021mnre}. %Existing multimodal learning models learn the correlations between text and the corresponding images by dual graph alignment~\cite{zheng2021multimodal, feng2023towards}, pairwise relation alignment~\cite{yuan2023joint, fu2022drake}, reinforcement learning-enhanced augmentation~\cite{xu2022different}, and various visual and textual feature fusion techniques~\cite{hu2023multimodal, li2023dual, chen2022good}. However, some of these current methods only consider whole-image visual features instead of features for just the objects in the image. Moreover, all existing work only focuses on multimodal relation extraction in supervised learning. 
Existing few-shot multi-modal fusion has been studied in the areas of visual question answering~\cite{tsimpoukelli2021multimodal, najdenkoska2023meta, jiang2023mewl}, image caption~\cite{alayrac2022flamingo, moor2023med}, action recognition~\cite{wanyan2023active, ni2022multimodal}, sentiment analysis~\cite{yang2022few}, and so on.
Studies have demonstrated that the performance of these tasks can be improved by fusing information from different modalities in few-shot learning~\cite{lin2023multimodality}.
Inspired by these works, we consider fusing visual information for few-shot relation extraction to provide the missing context in texts.
The only work on few-shot relation extraction focuses on social relation extraction, in which relations describe connections only between people~\cite{wan2021fl}. 
Besides, the dataset in~\cite{wan2021fl} is not in English and includes a limited number of classes, and is therefore not sufficient to conduct 10-way-K-shot learning experiments. 
Considering the above limitations, we focus on few-shot general relation extraction that is conducted on (1) a re-splitted MNRE dataset to satisfy few-shot learning, and (2) a subset of the FewRel dataset, where we collected corresponding images, to explore relation extraction in FSL.
%However, these fusion methods may bring semantic drift when the text and image are irrelevant.
%For relation extraction, relation happens between two name entities.
%Thus, the semantic information of the visual objects appearing in the images is also important.
%We propose image-guided attention, object-guided attention and hybrid feature attention to establish the interaction between pixel-level and object-level visual information with text to improve the few-shot relation extraction performance.

\section{Methodology}%Done

%In this section, we introduce the overview of MFS-HVE model.
Figure~\ref{fig:overview} shows the architecture of MFS-HVE for few-shot relation extraction. It consists of Semantic Feature Extractors and Multi-Modal Fusion. We describe these parts in detail below, starting with problem formulation.

\subsection{Problem Definition}
We follow the N-way-K-shot definition and settings of few-shot learning from~\cite{Gao_Han_Liu_Sun_2019} to conduct our experiments.
The N-way-K-shot setting means N classes with K examples of each.
Typically K is no more than 10.
There is no overlap between the classes in training data and testing data.
For the multi-modal few-shot relation extraction task, we tend to classify the relation between two name entities based on text and image inputs.
Let the input dataset represented by a set of tuples $(x_{i}, h_{i}, t_{i}, y_{i}, r_{i})$, where $x_{i}$ is a sentence, $h_{i}$ is a head entity, $t_{i}$ is a tail entity, $y_{i}$ is the corresponding image and $r_{i}$ is the relation between $h_{i}$ and $t_{i}$.
Our goal is to train a few-shot learning model $M$ to learn the representation function for the above tuples so that when randomly given support set with N relations and corresponding K tuples (NK tuples in total) as well as a query set with the same N relations and Q tuples, the model $M$ can predict the relations in the query set base on the given support set.
$M$ is learned by minimizing the semantic distance between the input embedding from the support set and the embedding from the query set.
At test time, we use a different set of relations and evaluate performance on the query set, given the support set.

\subsection{Semantic Feature Extractor}
Each instance contains a text message and a corresponding image for relation extraction.
The text is the input for the textual feature extractor, and the image is the input for the visual feature extractor.

\subsubsection{Textual Feature Extractor}\label{sec:textExtractor}
For the textual feature extractor, we use a pre-trained language model BERT~\cite{Devlin2019BERTPO} as the sentence encoder to generate the contextual representation.
Two unique tokens [CLS] and [SEP] are appended to the first and last positions.
The input text message is first tokenized into word pieces, and the positions of the name entities are marked by four special tokens [SEP] at the start and end of each entity mentioned in the relation statement of ~\cite{baldini-soares-etal-2019-matching}.  
Then output representation of the textual feature extractor $r_{t}$ can be formulated as follows:
\begin{equation}
    v_{i} = f_{\phi }(x_{i},h,t)
\end{equation}
\begin{equation}\label{equ:text}
    r_{t} = tanh(W \cdot v_{i} + b)
\end{equation}
where $v_{i}$ is the output of sentence encoder, $f_{\phi }$ is BERT encoder, $x_{i}$ is the input sentence, and $h$ and $t$ are head and tail entities, respectively.
A fully-connected layer is added after BERT encoder, where $W \in \mathbb{R}^{256\times 768}$ and $b \in \mathbb{R}^{256}$ are trainable.

\subsubsection{Visual Feature Extractor}~\label{sec:viExtractor}
\paragraph{Object Feature Representation}
Object-level features are considered as the semantic information of the objects appearing in the image instead of the features of the whole image.
For relation extraction tasks, a relation happens between the two name entities. 
Different from other multi-modal representation tasks, semantic information of the objects appearing in the images is of great importance.
To extract objects from images, we utilize the pre-trained object detection model Yolo~\cite{bochkovskiy2020yolov4} to recognize the objects in the images.
We consider the top $K$ frequent objects detected in the images to be the object labels because, in most cases, only the salient objects in the images are related to the name entities.
Then, we transform the object labels into object embeddings to augment the semantic information of the two name entities and address the problem of semantic disparity of different modalities.
The representation of object-level features $r_{o}$ can be expressed as:
\begin{equation}
    o = g_{\phi }(y_{i})
\end{equation}
\begin{equation}\label{equ:obj}
    r_{o} = f_{\phi }(o_{0})\oplus\cdots \oplus f_{\phi }(o_{k})
\end{equation}
where $g_{\phi }$ denotes the object detection model, $y_{i}$ is the input image, $\begin{Bmatrix}
o_{0}, o_{1}, \cdots , o_{k}
\end{Bmatrix} \in o$, indicating the objects detected in the image, $f_{\phi }$ is the object embedding encoder and $\oplus$ denotes concatenation.

\paragraph{Image Feature Representation}
The global image features are extracted from ResNet18~\cite{7780459}.
We use features from the last layer to produce the global vector.
We then transform each feature vector into a new vector with the same dimension as the representation of the textual features using a single-layer perception.
The representation of image-level features $r_{i}$ is:
\begin{equation}
    v_{i} = h_{\phi }(y_{i})
\end{equation}
\begin{equation}\label{equ:image}
    r_{i} = tanh(W \cdot v_{i} + b)
\end{equation}
where $h_{\phi }$ denotes the image encoder, $y_{i}$ is the input image, $W \in \mathbb{R}^{256\times 512}$ and $b \in \mathbb{R}^{256}$ are trainable weights and bias.

\subsection{Multi-Modal Fusion}~\label{sec:fusion}
\begin{figure}[htp] 
 \center{\includegraphics[height=5cm,width=7cm]{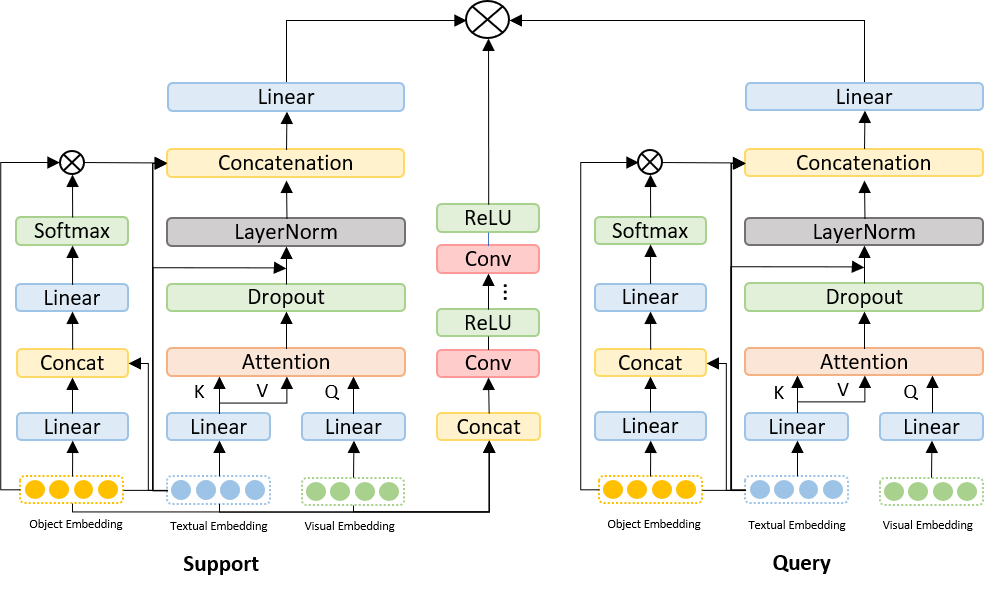}}
 \caption{\label{fig:fusion} Detailed structure of multi-modal fusion.}
 \end{figure}

%The multi-modal fusion module plays an important role in MFS-HVE. 
The architecture of our proposed multi-modal fusion is shown in Figure~\ref{fig:fusion}, including image-guided attention, object-guided attention, and feature-level attention.

\subsubsection{Image-Guided Attention}\label{sec:piAtt}
A cross-modal attention layer can provide a more sophisticated fusion between different modalities~\cite{sun-etal-2020-riva}.
Hence, we design a cross-attention layer module that combines the images and texts to capture the semantic interaction between visual regions of images and texts.
As shown in Figure~\ref{fig:fusion}, the cross-modal attention layer is image-guided attention, which is calculated by combining Key-Value pairs from one modality with the Query from another modality.
Specifically, the multi-modal representation is computed based on a modified version of the Scaled Dot-Product Attention (SA)~\cite{NIPS2017_3f5ee243}.
The attended feature for images $ \hat{f_{i}} = GA(q_{i},k_{t},v_{t})$ is obtained by reconstructing $q_{i}$ using all samples in $v_{t}$ for their normalized cross-modal similarity to $q_{i}$.
The image-guided attention unit is:
\begin{equation}
    GA(q, k, v) = softmax((\frac{(W_{Q}q)(W_{K}k)^{T}}{\sqrt{d_{k}}})W_{V}v)
\end{equation}
where $W_{Q}$, $W_{K}$, $W_{V}$ are trainable query, key and value parameters and $d_{k}$ is the dimension of key vectors.
Note that queries are from visual images, while keys and values are from text.

For each instance, the textual representation $r_{t} \in \mathbb{R}^{n \times d_{t}}$ is obtained through Equation~\ref{equ:text} and image representation is obtained through Equation~\ref{equ:image}. 
We first input the textual representation $r_{t}$ and image representation $r_{i}$ into fully connected layers, respectively.
Then, the image-guided attention unit models the pairwise relationship between the paired sample $<r_{t}, r_{i}>$, where $r_{i}$ guided the attention learning for $r_{t}$.
The new image-guided feature vector related to $r_{t}$ based on the cross-modal attention can be expressed as:
\begin{equation}
    \hat{r_{i}} = LayerNorm(r_{t} + GA(r_{i}, r_{t}, r_{t}))
\end{equation}
where LayerNorm is used to stabilize the training.

\subsubsection{Object-Guided Attention}\label{sec:obAtt}
Name entities in the textual sentence are always related to some objects detected from the input image.
As shown in Figure~\ref{fig:fusion}, we propose an object-guided attention unit to fuse relevant words (name entities) and visual regions (objects).
Given a textual feature $r_{t}$ obtained from Equation~\ref{equ:text} and a local object feature $r_{o}$ obtained from Equation~\ref{equ:obj}, we feed these features into a single neural network layer followed by a softmax function to generate the attention distribution over the objects:
\begin{equation}
    v_{r_{t}}=tanh(W_{r_{t}}r_{t}\oplus (W_{r_{o}}r_{o}+b_{r_{o}}))
\end{equation}
\begin{equation}
    \alpha _{r_{t}}=softmax(W_{a_{t}}v_{r_{t}}+b_{a_{t}})
\end{equation}
where $r_{t} \in \mathbb{R}^{d}$, $ r_{o} \in \mathbb{R}^{d}$, $W_{r_{o}}$, $W_{r_{t}}$, $W_{a_{t}}$, $b_{r_{t}}$ and $b_{a_{t}}$ are all trainable weights and bias.
$\oplus$ denotes concatenation.
Based on the attention distribution $a_{t}$, the new object vector $\hat{r_{o}}$ related to $r_{t}$ is:
\begin{equation}
    \hat{r_{o}}=\sum \alpha _{r_{t}}r_{o}
\end{equation}

\subsubsection{Hybrid Feature Attention}~\label{sec:feaAtt}
As shown in the middle of Figure~\ref{fig:fusion}, the hybrid feature attention fuses text information, global image-guided visual information, and local object-guided information, highlighting the important dimensions in the joint feature space to alleviate feature sparsity.
For few-shot relation extraction, only a few instances in the support set are used for training so that the features extracted from the support set suffer from the problem of data sparsity.
The feature-level attention generation block contains one concatenation layer, two or three 2D convolutional layers, and two or three activation functions, which can pay more attention to those more discriminative features when computing the space distance.

For space distance, studies show that hyperbolic spaces, where suitable curvatures match the characteristics of data, can lead to more generic embedding spaces~\cite{9711172, Liu_2020_CVPR}.
In the example shown in Figure~\ref{fig:example}, the detected object `person' is the hypernym of the name entity `Magic Johnson' in the text.
Thus, we adopt hyperbolic distance with feature-level attention in our networks to preserve such hierarchical structure:
\begin{equation}\label{equ:distance}
    d(s_{1}, s_{2}) = \alpha_{i}\cdot cosh^{-1}(1+2\frac{\left \| s_{1}-s_{2} \right \|^{2}}{(1-\left \| s_{1} \right \|^{2})(1-\left \| s_{2} \right \|^{2})})
\end{equation}
where $\alpha_{i}$ is the score vector for relation $r_{i}$ calculated via the hybrid feature attention shown in Figure~\ref{fig:fusion}.
By multiplying the hybrid feature attention weight by the support and query embeddings, we make the distance metrics better fit the given support sets and relations.

\subsection{Model Training}

The objective of training MFS-HVE is to minimize the distance between each instance embedding $L_{multi}$ and the relation embedding $P_{multi}(S)$.
A cross-modality encoder concatenates the three vectors: sentence embedding $r_{t}$, object-guided textual embedding $\hat{r_{o}}$, and image-guided textual embedding $\hat{r_{i}}$, to yield the multi-modal representation.
Then, a fully connected layer is added to refine the multi-modal representation.
The final multi-modal instance embedding $L_{multi}$ is:
\begin{equation}
    L_{multi} = tanh(W_{multi} \cdot (r_{t} \oplus \hat{r_{o}} \oplus \hat{r_{i}}) + b_{multi})
\end{equation}
where $W_{multi}$ and $b_{multi}$ are trainable.

Given support set $S$ in the $N$ way $K$ shot setting, we compute a prototype for each of the $N$ relations $R$ in $S$ based on the multi-modal representations $L_{multi}$ of $K$ tuples.
To be more specific, the prototype representation $P_{multi}(S)$ for $R$ is shown as:
\begin{equation}
    P_{multi}(S) = \frac{1}{K}\sum_{i=1}^{K}L_{multi}
\end{equation}
To predict the final relation among $N$ ways, hyperbolic distance $d$ as shown in Equation~\ref{equ:distance} is calculated between a query instance and each prototype $P_{multi}(S)$.
Then, a softmax function is applied over the distance vector to generate a probability distribution on relations.
More precisely, the probabilities of the relations for a query instance $q$ are computed as:
\begin{equation}
    Pr(y=r_{i}|q) = \frac{exp(-d((L_{multi}),P_{m}(S)))}{\sum_{i=1}^{\left | R \right |}exp(-d((L_{multi}),P_{i}(S)))}
\end{equation}
where $d(\cdot)$ is the hyperbolic distance.

\section{Experiments}
We conducted experiments with ablation studies, case studies, and parameter sensitivity experiments on two public datasets: MNRE and FewRel, to show that integrating semantic visual information with object-level and global feature-level attention mechanisms can help improve the performance.

\subsection{Datasets}~\label{sec:dataset}
\begin{table}[!htb]
\small
\caption{The statistics of each dataset.}
\label{tab:dataset}
\begin{tabular}{cccc}
\hline
         & \#instances & \#relations & avg. len. \\ \hline
MNRE      & 15,484     & 23          & 16.67          \\
FewRel   & 56,000      & 80          & 24.95     \\
FewRel\textsubscript{small} & 3,703      & 80         & 23.90     \\
 \hline
\end{tabular}
\end{table}
In our experiments, we evaluate our model~\footnote{Code is available: \url{https://github.com/gjiaying/MFS-HVE}} over two widely used datasets: MNRE~\cite{10.1145/3474085.3476968}, FewRel~\cite{han-etal-2018-fewrel}, and a subset of FewRel, which includes only clean images. 
FewRel is a balanced dataset, and MNRE is an unbalanced dataset. 
The statistics of MNRE and FewRel datasets are shown in Table~\ref{tab:dataset}.
We describe each dataset and dataset construction in detail in Appendix~\ref{sec:A1}.
For the MNRE dataset, we randomly re-split the original supervised MNRE dataset to ensure that there is no overlap of relations between the training set and testing set. 
For FewRel and FewRel\textsubscript{small} datasets, we follow the same training and validation set.
%Details of hyperparameters and configuration settings are discussed in Sec.~\ref{sec:A2} and Table~\ref{tab:parameter}.

\subsection{Baselines and Evaluation Metrics}~\label{sec:baselines}
%Baselines
We compare our model with six only text-based models: \textbf{Siamese}~\cite{koch2015siamese}, \textbf{Proto}~\cite{NIPS2017_cb8da676},  \textbf{SNAIL}~\cite{mishra2018a},  \textbf{GNN}~\cite{garcia2018fewshot}, \textbf{MLMAN}~\cite{ye-ling-2019-multi},
 \textbf{MTB}~\cite{baldini-soares-etal-2019-matching} and eight text-based models with external information:  \textbf{REGRAB}~\cite{qu2020few},  \textbf{ZSLRC}~\cite{10.1145/3459637.3482403},  \textbf{ConceptFERE}~\cite{yang-etal-2021-entity}, \textbf{MapRE}~\cite{dong-etal-2021-mapre}, \textbf{HCPR}~\cite{han2021exploring}, 
 \textbf{GM\_GEN}~\cite{li-qian-2022-graph},
 \textbf{FAEA}~\cite{ijcai2022p407} and \textbf{SimpleFSRE}~\cite{liu-etal-2022-simple}.
For multi-modal fusion baselines, we considering fusing the information from different modalities at different levels.
The early fusion includes \textbf{Concatenation}~\cite{wan2021fl}, and \textbf{Circulant Fusion}~\cite{gong2023circulant}.
The mid-level fusion includes \textbf{Deep Fusion}~\cite{wang2020deep},  \textbf{Dual Co-Att}~\cite{liu2021dual}, and \textbf{Proto\textsubscript{multimodal}}~\cite{ni2022multimodal}.
%Evaluation
We follow the same settings as~\cite{qu2020few} to run the experiments.
The evaluation metric is the Accuracy (Acc.) of query instances.

\subsection{Parameter Settings}
\label{sec:A2}
\begin{table}[!htb]
\caption{Parameter Settings}
\label{tab:parameter}
\centering
\begin{tabular}{lr}
\hline
Parameter                            & Value     \\ \hline
Textual Information Dimension $d_{t}$     & 512        \\
Visual Information Dimension $d_{v}$ & 128         \\
Object Information Dimension $d_{o}$  & 256 \\
Batch Size                           & 1         \\
Initial Learning Rate $\alpha$       & 0.1      \\
Weight Decay                         & $10^{-5}$ \\
Dropout                       & 0.2    \\ 
Sentence Max Length            &128   \\
Objects Number                 & 2   \\
\hline

\end{tabular}
\end{table}

For the hyperparameter and configuration of MFS-HVE, we implement MFS-HVE based on the PyTorch framework and optimize it with AdamW optimizer.
We report the result based on a five-times run of the experiment. GPU of 16G memory is needed for the training process. 
The training time is around 5-6 hours depending on the computing resource.
For the sentence encoder, we initialize the textual representation by pre-trained BERT~\cite{Devlin2019BERTPO} and set the dimension size at 768. 
Then we follow ~\cite{baldini-soares-etal-2019-matching} to combine the token encodings of the entity mentioned in the sentence.
For the image encoder, we initialize the visual representation by pre-trained ResNet18~\cite{7780459} and set the dimension size at 512.
For the object encoder, we employ 50-dimensional GloVe (6B tokens, 400K vocabulary)~\cite{pennington2014glove} for word embeddings of the objects detected from the image.
Table~\ref{tab:parameter} shows other parameters used in the experiment.

\subsection{Results and Discussion}~\label{sec:experiment}

\begin{table*}[]
\small
\caption{Results of Accuracy Comparison Among Models (\%) on MNRE and FewRel\textsubscript{small} Datasets. }
\label{tab:mainresult}
\centering
\tabcolsep=0.18cm
\begin{tabular}{c|l|cc|cccc}
\hline
                                                                &                         & \multicolumn{2}{c|}{MNRE}       & \multicolumn{4}{c}{FewRel\textsubscript{small}}                            \\ \cline{3-8} 
                                                                &                         & 5-Way          & 10-Way         & 5-Way          & 5-Way          & 10-Way & 10-Way         \\
\multirow{-3}{*}{Modality}                                      & \multirow{-3}{*}{Model} & 1-Shot         & 1-Shot         & 1-Shot         & 5-Shot         & 1-Shot & 5-Shot         \\ \hline
                                                                & GNN~\cite{garcia2018fewshot}                     & 29.08          & 22.53          & 46.38          & 70.45          & 28.74  & 62.07          \\
                                                                & Snail~\cite{mishra2018a}                  & 30.90          & 19.43          & 40.16          & 60.07          & 21.19  & 47.56          \\
                                                                & Siamese~\cite{koch2015siamese}                 & 36.08          & 26.50          & 62.74          & 73.92          & 42.17  & 65.05          \\
                                                                & MLMAN~\cite{ye-ling-2019-multi}                     & 35.08          & 29.06          & 63.47          & 74.47          & 61.86  & 72.58          \\
                                                                & Proto\_BERT~\cite{NIPS2017_cb8da676}             & 49.75          & 33.57          & 75.64          & 84.64          & 64.17  & 75.27          \\
\multirow{-6}{*}{Only Text}                                     & MTB~\cite{baldini-soares-etal-2019-matching}                     & 46.02          & 32.35          & 76.38          & 86.27          & 65.27  & 73.81          \\ \hline
                                         & ZSLRC~\cite{10.1145/3459637.3482403}              & 45.65          & 32.23          & 71.82          & 81.74          & 64.88  & 71.81          \\
                                        
                                        & ConceptFERE~\cite{yang-etal-2021-entity}                    & -          & -          & 75.86          & 83.38          & 68.38  & 76.06\\
                                        & REGRAB~\cite{qu2020few}                    & -          & -          & 78.53          & 84.96          & 70.65  & 78.00\\
                                        & HCRP~\cite{han2021exploring}                    & 31.10          & 10.45          & 78.04          & 84.68          & 69.54  & 77.91          \\
                                        & MapRE~\cite{dong-etal-2021-mapre}                   & 51.92          & 35.20          & 79.44          & 85.60          & 70.71  & 78.84         \\
                                        & GM\_GEN~\cite{li-qian-2022-graph}                  & 52.58          & 35.82          & 60.04          & 73.74          & 42.22  & 59.23          \\
                                        & FAEA~\cite{ijcai2022p407}                   & 52.14          & 33.37          & 80.80          & 87.94          & 71.30  & 79.29  \\
\multirow{-8}{*}{Text+Others} & SimpleFSRE~\cite{liu-etal-2022-simple}                      & 50.32          & 35.05          & 80.84           & 87.46           & \textbf{71.67}   & 80.14           \\ \hline
                                                                & Concat~\cite{wan2021fl}                  & 40.17          & 29.83          & 74.10          & 84.69          & 66.08  & 75.95          \\
                                                                & CirculantFusion~\cite{gong2023circulant}                     & 38.39          & 29.19          & 73.21          & 83.58          & 65.11  & 76.29          \\
                                                                & DeepFusion~\cite{wang2020deep}              & 48.27          & 33.28          & 78.38          & 86.76          & 66.36  & 76.08          \\
                                                                & Proto\textsubscript{multimodal}~\cite{ni2022multimodal}                    & 50.84          & 34.10          & 77.18          & 86.28          & 68.19  & 78.29          \\
                                                                & Dual Co-Att~\cite{liu2021dual}                    & 52.52          & 35.62          & 77.60          & 87.24          & 68.69  & 78.54          \\ \cline{2-8} 
\multirow{-6}{*}{Text+Image}                                    & \textbf{MFS-HVE}        & \textbf{54.88} & \textbf{36.62} & \textbf{81.32} & \textbf{89.65} & 69.52  & \textbf{80.55} \\ \hline
\end{tabular}
\end{table*}

\subsubsection{Main Results}~\label{sec:fewresult}
The experiment results of few-shot learning on MNRE and FewRel\textsubscript{small} are shown in Table~\ref{tab:mainresult} with the average of five times run.
Because some relations have less than ten instances in MNRE, it is impossible to run 5-shot experiments on MNRE. because we need five instances for the support set and the same number of instances for the query set. 
Thus, we only run 1-shot experiments on MNRE.
FewRel is a public dataset with only textual information.
We crawl the image relevant to each textual instance to construct a few-shot multi-modal dataset: FewRel\textsubscript{small}, which is a subset of FewRel, including only clean images.
Note that the baselines of multi-modal fusion works are implemented based on MTB~\cite{baldini-soares-etal-2019-matching} to have a fair comparison in few-shot relation extraction.

From Table~\ref{tab:mainresult}, we observe that models integrating external information (labels, graphs, images, etc) perform much better than only text-based models.
Models fusing semantic visual information can help improve the performance, but the performance highly depends on the fusion methods.
Simply concatenating the visual information or fusing information at a coarse-grained level without considering semantic meanings such as circulant fusion may negatively impact the performance.
%Fusing information at a coarse-grained level without semantic meanings such as circulant fusion may also negatively impact the performance.
This is probably because these methods treat all visual and textual information with equal importance (weights). 
However, only partial visual images contain relevant semantic meanings to the text.
Directly using all information in the image may bring noise to the textual data.
%After considering image-guided textual information, object-guided textual information, and joint learning of text, image, and objects, MFS-HVE outperforms all other models.
We further explore the robustness in Appendix~\ref{sec:robust}.
After considering image-guided textual information, object-guided textual information, and joint learning of text, image, and objects, our proposed model MFS-HVE significantly outperforms all state-of-the-art models on MNRE.
More details about the performance of different attention layers of MFS-HVE are discussed in the ablation study in Section ~\ref{sec:ablation}.

%\paragraph{Results on FewRel}
%From Table~\ref{tab:fewrelsmall}, we find that models based on multi-modal information perform better than text-based models in general.
%However, not all multi-modal models can achieve a better result.
%Simple concatenation or circulant multiplication (MCF) of information from different modalities even has worse performance than text-based models such as MTB.
%We think that such performance decrease is caused by fusion without considering semantic contexts in the image.
%From table~\ref{tab:fewrelsmall}, we can observe that our proposed model MFS-HVE, which integrates semantic textual and visual information at both global and local levels, achieves the best performance among all models.

%\begin{comment}

%paragraph{Results Summary}
In summary, based on the experiment results on MNRE, FewRel, and FewRel\textsubscript{small}, we have the following findings:
\begin{enumerate}
    \item We find that models with multi-modal information perform better than text-based models in general.
    \item Multi-modal models based on high-quality visual information are more robust than text-based models when the dataset size becomes smaller.
    \item The performance of multi-modal models highly depends on the fusion methods. Simple concatenation or circulant multiplication of information from different modalities may probably have a negative impact.
    \item For the relation extraction task, the local object information from the image is also very important because they are related to name entities in textual sentences and help reduce the noise of global image features.
\end{enumerate}

%\end{comment}

\subsubsection{Ablation Study}\label{sec:ablation}%Done

\begin{table*}[]
\caption{Ablation study over MFS-HVE components (\%) on MNRE and FewRel\textsubscript{small} datasets.}
\label{tab:ablation}
\centering
\begin{tabular}{l|cc|cccc}
\hline
\multirow{3}{*}{Model Component} & \multicolumn{2}{c|}{MNRE}       & \multicolumn{4}{c}{FewRel\textsubscript{small}}                                    \\ \cline{2-7} 
                                 & 5-Way          & 10-Way         & 5-Way          & 5-Way          & 10-Way         & 10-Way         \\
                                 & 1-Shot         & 1-Shot         & 1-Shot         & 5-Shot         & 1-Shot         & 5-Shot         \\ \hline
Only Text                        & 49.39          & 31.95          & 76.66          & 85.82          & 63.54          & 76.73          \\
Image Attention                  & 50.43          & 32.40          & 78.37          & 86.75          & 66.28          & 77.18          \\
Object Attention                 & 50.57          & 33.63          & 78.85          & 86.24          & 66.96          & 77.96          \\
Image\&Object Attention          & 52.26          & 35.38          & 80.50          & 88.72          & 69.49          & 79.17          \\
\textbf{MFS-HVE}                 & \textbf{54.88} & \textbf{36.62} & \textbf{81.32} & \textbf{89.65} & \textbf{69.52} & \textbf{80.55} \\ \hline
\end{tabular}
\end{table*}

\begin{figure*}[htp] 
 \center{\includegraphics[height=7.5cm,width=\textwidth]{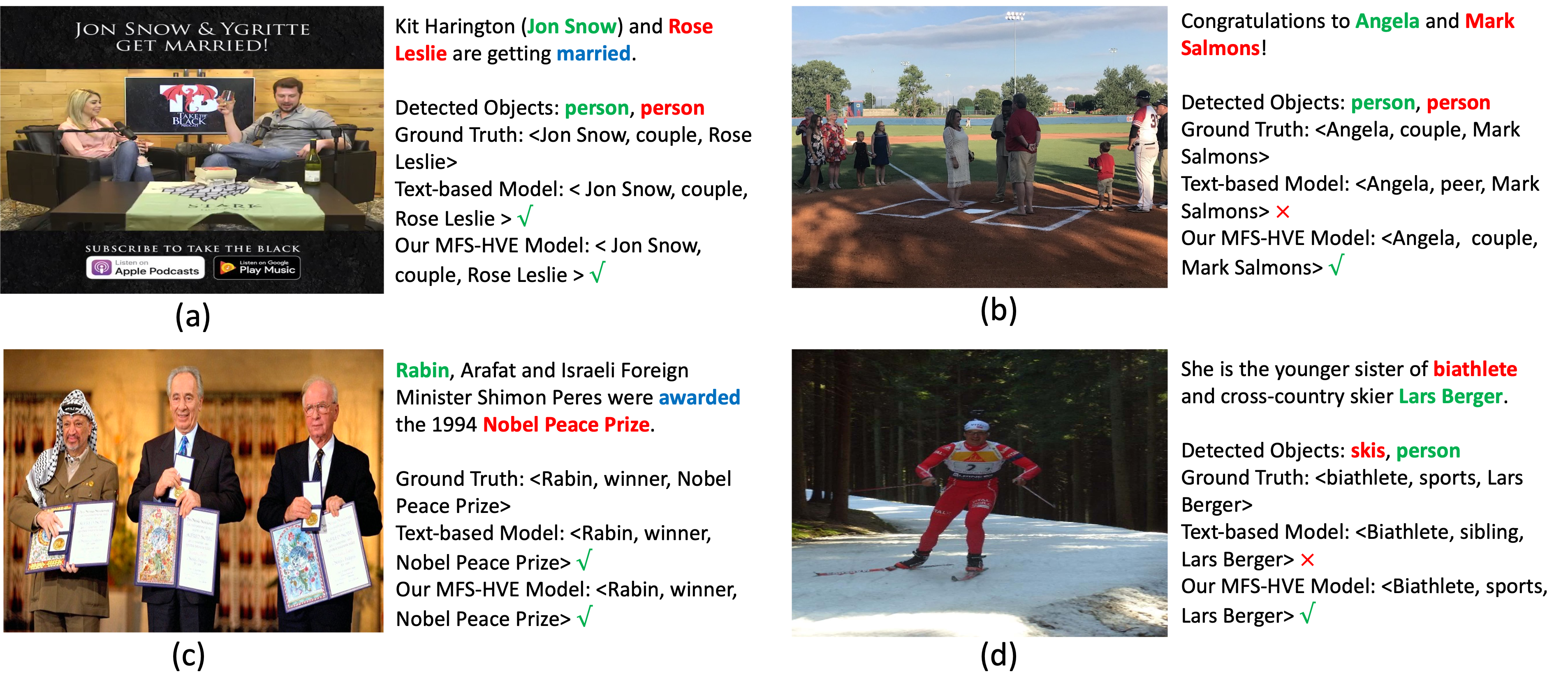}}
 \caption{\label{fig:case} The examples of our proposed model MFS-HVE comparing to a text-based model on both the MNRE and FewRel datasets. We present the relation extraction results with the detected objects from the relevant image in the right column. The head entities are highlighted in green, whereas the tail entities are highlighted in red.}
 \end{figure*} 

To illustrate the effectiveness of MFS-HVE and explore the role of each attention unit in MFS-HVE, we carry out the ablation study on the datasets only with clean and high-quality visual data (MNRE and FewRel\textsubscript{small}) because the performance of fusion with different multi-modal information is unstable with noisy data.% (Appendix~\ref{sec:robust}).
%We further explore the usage of different attention units 
The ablation experiment results shown in Table~\ref{tab:ablation} are reported by the mean value of five times the experimental results. We observe that utilizing multi-modal information performs better than uni-modal information (text).
However, only using image-guided attention or object-guided attention can not achieve a great performance improvement.
This is probably because considering the whole image from a global perspective may introduce noise to the text, resulting in a similar performance in few-shot settings compared with text-based models.
In addition, if only object-guided textual attention is added to the model, the model still can not achieve a significant improvement.
This is because not all images include the objects that are relevant to the name entities in the text.
Thus, when the model jointly fuses image attention and object attention, there is a promising performance increase.
The image attention overcomes the problem of sparsity, whereas the object attention reduces the noise brought by the whole image features.
After adding hybrid feature attention to fuse all textual and visual information from both global and local perspectives, a significant performance gain is seen.

\subsubsection{Case Study}%Done
Figure~\ref{fig:case} shows the case study comparing our MFS-HVE model with a text-based model MTB on both MNRE and FewRel datasets.
To evaluate the advantage and effectiveness of semantic visual information, we compare our model with an unimodal model, which only depends on textual information.
We present four examples of two relations.
For each relation, we present two cases.
One case is that both the text-based model and the multimodal model MFS-HVE predict the relation correctly.
The other case is that the relation is incorrectly predicted by the text-based model but correctly predicted by MFS-HVE.

Based on these examples, we observe that the text-based model only performs well when rich information is in the text.
For the examples shown on the left, the text-based model can only correctly predict the relation `couple' when relevant words or phrases with similar meanings appear in the text, such as `married' in the first sentence.
Similarly, for the relation `winner', the text-based model also performs well when the long textual sentence contains detailed information such as the word `awarded'.
These words relevant to the target relations provide enough semantic hints for the models with only text.
However, not all cases have such long or detailed textual hints for the model.
In the examples shown on the right, the textual sentences are short, without any words related to the target relation.
In these cases, the text-based model can not predict the relation correctly.
The text-based model predicts `Angel' and `Mark' are peers instead of `couple', `Roger Federer' is the `participant of' the tennis tournament `Wimbledon' instead of `winner' of `Wimbledon'.
Nevertheless, with the guidance of informative visual evidence, more semantics are provided to the text.
In the upper-right example, a wedding ceremony is shown in the image, and people objects are detected in the image.
Based on this information, MFS-HVE correctly predicts the relation `couple' instead of other relations in the MNRE dataset such as `sibling', `peer', `parent', etc.
Similarly, in the lower right example, MFS-HVE predicts the relation `Roger Federer' is the `winner' of `Wimbledon' based on the visual information that a person is holding a tennis racket.
In summary, integrating semantic visual information at both global and local levels provides more relevant information to supplement the missing contexts in textual sentences, resulting in a better and more robust performance for few-shot relation extraction.

%Figure~\ref{fig:case} shows the case study comparing MFS-HVE with a text-based model MTB on both MNRE and FewRel datasets.
%To evaluate the effectiveness of semantic visual information, we compare our model with an MTB, which only depends on textual information.
%We observe that the text-based model only performs well when rich information is in the text.
%In the examples shown on the right, the textual sentences are short, without any words related to the target relation.
%Thus, the text-based model can not predict the relation correctly.
%In (b), a wedding ceremony is shown in the image, and people objects are detected in the image.
%Based on this information, MFS-HVE correctly predicts the relation `couple' instead of other relations in the MNRE dataset such as `peer', `parent', etc.
%In (d), MFS-HVE predicts the relation `biathlete' is the `sports' of `Lars Berger' based on the visual information that a person is on skis. 
%The word `sister' is noisy information resulting in a false prediction of the relation `sibling' without visual information.

\subsubsection{Parameter Sensitivity}
\begin{figure}[htp] 
 \center{\includegraphics[height=6cm,width=7cm]{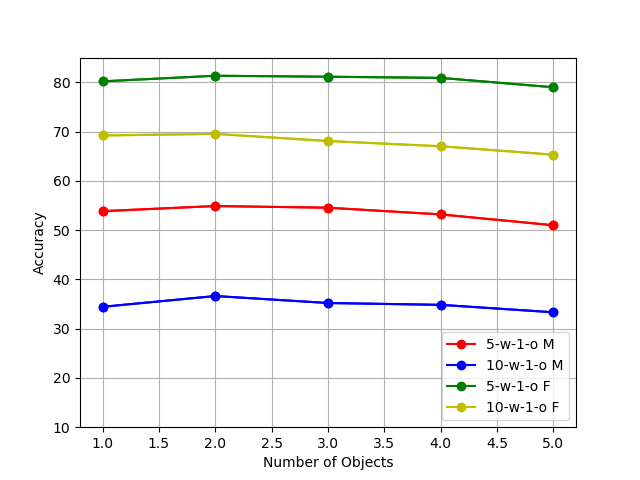}}
 \caption{\label{fig:objects} Effects on varying the number of embedded objects in one-shot settings on MNRE and FewRel\textsubscript{small} datasets.} 
\end{figure}

Figure~\ref{fig:objects} shows the results of our proposed MFS-HVE model influenced by embedding a different number of objects detected from the image.
By varying the object number from one to five, the results in terms of Accuracy on both MNRE and FewRel\textsubscript{small} are exhibited in Figure~\ref{fig:objects}.
We observe that the object number affects the performance of few-shot relation extraction.
The model achieves the best performance when the object number is two.
The performance drops when the object number increases.
This is reasonable because relations always happen between two name entities.
The two detected objects are usually relevant to the two corresponding name entities if the images are of high quality.
Embedding only one object may lose critical information, whereas embedding lots more objects also introduces noise (irrelevant information) to the visual information.

\section{Conclusion and Future Work}%Done
In this paper, we propose MFS-HVE, a multi-modal few-shot relation extraction approach leveraging semantic visual information to supplement the missing contexts in textual sentences.
Our multi-modal fusion module consists of image-guided attention, object-guided attention, and hybrid feature attention that integrates information from different modalities.
Experimental results demonstrate that MFS-HVE leveraging attention-based multi-modal information outperforms other uni-modal baselines and multi-modal fusion methods in few-shot relation extraction.
In future work: (1) We will implement other powerful SOTA image encoders such as ViT~\cite{DBLP:conf/iclr/DosovitskiyB0WZ21} to generate feature-level image embeddings. (2) We will explore utilizing the semantic visual information as an external source in zero-shot learning.

\section{Ethical Considerations}
\paragraph{Dataset Construction}
Because FewRel is a uni-modal relation extraction dataset, we obtain the images for each sentence from Wikidata, which is a free and collaborative knowledge base. Wikidata follows open data principles, which means that the data it contains is available to the public for various purposes, including research. We collect instance-related images from the wiki to make a multi-modal relation extraction dataset for experiments. Because not all instances from FewRel have relevant images, we remove the instances that do not have a corresponding relevant image on FewRel, resulting in a new dataset FewRel\textsubscript{small}

\paragraph{Computing Cost} 
Our proposed model MFS-HVE requires GPU training, which imposes a computational burden. Specifically, our model needs 5-6 hours of training on a single GPU card, which results in 0.25lbs of carbon dioxide emissions.

%Our model needs the use of GPU training, which imposes a computational burden. We acknowledge that this burden has an adverse environmental impact in terms of carbon emissions. Specifically, our research requires 6 hours training on a single GPU card for each task. In total, we have 6 $\times$ 5 (each task has 5 runs) $\times$ 15 tasks (6 tasks in RTE, 4 tasks in RC and 5 tasks in ablation study) = 450 hours training, resulting in 112.5lbs of carbon dioxide.

%\bibliographystyle{acl_natbib}
\nocite{*}
\section{Bibliographical References}\label{sec:reference}

\bibliographystyle{lrec-coling2024-natbib}
\bibliography{lrec-coling2024-example}

%\bibliographystylelanguageresource{lrec-coling2024-natbib}
%\bibliographylanguageresource{languageresource}
%\appendix

\section{Appendix}
\label{sec:appendix}

\subsection{Dataset Construction and Description}
\label{sec:A1}

In the following, we describe each dataset in detail:
\begin{itemize}
\item \textbf{\textit{MNRE~\cite{10.1145/3474085.3476968}.}}
The MNRE dataset is a public human-annotated unbalanced multi-modal neural relation extraction dataset.
It is originally built upon Twitter15~\cite{lu-etal-2018-visual}, Twitter17~\cite{Zhang_Fu_Liu_Huang_2018} and crawling data from Twitter~\footnote{https://archive.org/details/twitterstream}.
Each piece of data includes a sentence with two name entities and an image ID to correlate the text with the image.
Because MNRE is a relation extraction dataset for supervised learning, there is an overlap of relations between the training and the testing dataset. 
For few-shot relation extraction, we randomly re-split the MNRE dataset to ensure no overlap of classes between the training and testing sets.
There are 23 classes in total. After splitting the dataset, there are 13 classes for training and 10 classes for testing.

\item \textbf{\textit{FewRel~\cite{han-etal-2018-fewrel}.}}
The FewRel dataset is a public human-annotated balanced few-shot RC dataset consisting of 80 types of relations (64 for training and 16 for validation, another 20 for testing but it is not public), each of which has 700 instances. 
Because we need to combine images with the original text, so we only run experiments on the public part (64 training + 16 validation). 
Because FewRel is a fully uni-modal dataset, we insert an image ID to each instance to make it into a multi-modal relation extraction dataset.
The image for each instance is automatically crawled by a built-in web crawler~\footnote{https://github.com/hellock/icrawler} on wiki data from the Google search engine.

\item \textbf{\textit{FewRel\textsubscript{small}.}}
FewRel\textsubscript{small} is a subset of FewRel.
Because FewRel doesn’t have image information, we crawl the images for FewRel. 
We view these images as external information, similar to auxiliary information such as label description, knowledge graphs, entity description, etc.
Because images crawled for FewRel is an automatic process, some of the images are not relevant to their corresponding texts. 
Noise exists in the newly constructed multi-modal FewRel dataset.
Noisy images are removed to ensure that FewRel\textsubscript{small} is a small, clean, and high-quality multi-modal few-shot relation extraction dataset.
Note that we did not do any labeling work.
The labels remain the same in FewRel\textsubscript{small} as FewRel, and we only add more information (images) for the existing dataset.
\end{itemize}

In all, FewRel is a balanced dataset.
Due to the data cleaning, FewRel\textsubscript{small} is an unbalanced dataset.
MNRE is also an unbalanced dataset.

\subsection{Model Robustness}
\label{sec:robust}
To further study the robustness of integrating visual information with textual information, we also conduct experiments on the model's performance comparison on FewRel and FewRel\textsubscript{small}.
To make fair comparisons, instead of directly reporting the performance of other state-of-the-art models, we re-implement other models with the same parameter settings as the models run on FewRel\textsubscript{small}. 
Table~\ref{tab:robust} shows the results of performance decrease from dataset FewRel to FewRel\textsubscript{small} in few-shot settings.
Because the FewRel dataset is more than ten times larger than FewRel\textsubscript{small}, there are more training instances in FewRel.
It is reasonable to expect a performance drop when the model is training on a smaller dataset.
From Table\ref{tab:robust}, we observe that the performance of text-based models drops significantly when the dataset tends to be smaller.
This is because models usually can perform better when more data is available.
In addition, we also find that models based on multi-modal information are more robust than text-based models.
They have a smaller performance decrease than text-based models.
Our proposed model MFS-HVE performs the best in the one-shot learning setting.
We conjecture that the high-quality semantic visual information neutralizes the negative impact of little training data in FewRel\textsubscript{small}, resulting in a more robust performance of multi-modal models.

% \usepackage{multirow}
\begin{table*}[]
\caption{Results of performance decrease in Accuracy(\%) from
FewRel to FewRel\textsubscript{small}. }
\label{tab:robust}
\centering
\begin{tabular}{lcccc}
\hline
\multirow{2}{*}{Model} & 5-Way         & 5-Way         & 10-Way        & 10-Way        \\
                       & 1-Shot        & 5-Shot        & 1-Shot        & 5-Shot        \\ \hline
GNN~\cite{garcia2018fewshot}                    & 12.32         & 10.90         & 12.18         & 6.53          \\
Snail~\cite{mishra2018a}                  & 9.88          & 12.12         & 11.19         & 11.58         \\
Siamese~\cite{koch2015siamese}                & 5.61          & 8.36          & 14.24         & 4.25          \\
MLMAN~\cite{ye-ling-2019-multi}                  & 3.30          & 1.97          & 2.70          & 2.25          \\
Proto\_BERT~\cite{NIPS2017_cb8da676}            & 2.20          & 4.31          & 3.11          & 7.35          \\
MTB~\cite{baldini-soares-etal-2019-matching}                    & 3.14          & 1.00          & 3.54          & 3.66          \\
ZSLRC~\cite{10.1145/3459637.3482403}                    & 4.01          & 6.10          & 2.34          & 5.83          \\
ConceptFERE~\cite{yang-etal-2021-entity}            & 3.56          & 2.96          & 3.34          & 3.76          \\
REGRAB~\cite{qu2020few}                 & 4.32          & 4.88          & 3.44          & 4.07          \\
HCRP~\cite{han2021exploring}            & 4.36          & 3.00          & 2.76          & 4.24          \\
MapRE~\cite{dong-etal-2021-mapre}           & 6.29          & 7.24          & 8.47          & 8.80          \\
FAEA~\cite{ijcai2022p407}           & 7.97          & 6.78          & 4.55          & 5.59          \\
SimpleFSRE~\cite{liu-etal-2022-simple}          & 5.45          & 7.45          & 5.79          & 7.54          \\
Concat~\cite{wan2021fl}                 & 3.08          & 1.32          & 2.58          & 0.83          \\
DeepFusion~\cite{wang2020deep}            & 2.14          & 4.72          & 0.38          & \textbf{0.39} \\
CirculantFusion~\cite{gong2023circulant}                    & 3.99          & 2.60          & 5.75          & 2.24          \\
Dual Co-Att~\cite{liu2021dual}                  & 2.60          & 1.58          & 3.67          & 2.02          \\
Proto\textsubscript{multimodal}~\cite{ni2022multimodal}                   & 2.01          & 3.08          & 3.75          & 2.99          \\ \hline
\textbf{MFS-HVE}       & \textbf{1.95} & \textbf{0.83} & \textbf{0.27} & 1.32          \\ \hline
\end{tabular}
\end{table*}

\subsection{Limitations}
We view the following current limitations as some opportunities to build on in future work. First, MFS-HVE requires high-quality images for training. As shown in Table~\ref{tab:mainresult}, MFS-HVE has a significant performance improvement compared with models using other text-based external information on MNRE. This is because MNRE is a public multi-modal dataset including clean and high-quality images. However, MFS-HVE shows a slight improvement or similar performance with models using other text-based external information on FewRel. The images crawled automatically contain much noise, which means some of the crawled images are irrelevant to the textual sentences. To further improve the performance on the FewRel dataset, human efforts or other crawling techniques are needed to get a large, clean, and high-quality image dataset. 

Second, we compare MFS-HVE with five different fusion models introduced in Sec.~\ref{sec:related}. There are no existing multi-modal fusion models for the few-shot relation extraction task. We follow the five models' papers to implement the multi-modal fusion algorithms.
To meet the requirement for few-shot learning, these fusion methods are built upon MTB~\cite{baldini-soares-etal-2019-matching}. More latest multi-modal fusion methods are needed for performance comparison. To further improve the performance, more SOTA visual encoders such as ViT~\cite{DBLP:conf/iclr/DosovitskiyB0WZ21} and large GPU memories are needed to conduct more experiments.

Finally, we want to clarify that our work focuses on few-shot relation extraction. 
We compare our model's performance with 14 SOTA open-code few-shot RE models and 5 different fusion models on two public English datasets. State-of-the-art multi-modal models in supervised learning for other tasks (i.e. NER, etc) or other languages besides English, are outside the scope of our paper because not all supervised models could be adapted/changed to few-shot settings as the training process is completely different.
\end{document}